\documentclass[letterpaper, 10 pt, conference]{ieeeconf}  

\IEEEoverridecommandlockouts                              

\usepackage{amsmath,amsthm,amssymb}
\usepackage{mathtools}
\mathtoolsset{showonlyrefs}
\usepackage{mathrsfs}
\usepackage{comment}
\usepackage[caption=false,font=footnotesize]{subfig}
\usepackage{multicol,multirow}
\usepackage[dvipsnames,table]{xcolor}

\usepackage[ruled,linesnumbered]{algorithm2e}

    \makeatletter
    \let\NAT@parse\undefined
    \makeatother

\usepackage[hidelinks]{hyperref}

\DeclareMathOperator*{\argmax}{arg\,max}
\DeclareMathOperator*{\argmin}{arg\,min}

\newcommand{\bbR}{\mathbb{R}}

\newcommand{\tr}{^\intercal}

\newcommand{\bfy}{\mathbf{y}}
\newcommand{\bfx}{\mathbf{x}}

\newcommand{\bfk}{\mathbf{k}}
\newcommand{\bfK}{\mathbf{K}}
\newcommand{\bfI}{\mathbf{I}}

\LinesNumbered

\title{\LARGE \bf
Decentralized Learning With Limited Communications for Multi-robot Coverage of Unknown Spatial Fields 
}

\author{Kensuke Nakamura$^*$, María Santos$^*$, and Naomi Ehrich Leonard
\thanks{©2022 IEEE. Personal use of this material is permitted. Permission from
IEEE must be obtained for all other uses, in any current or future media,
including reprinting/republishing this material for advertising or promotional
purposes, creating new collective works, for resale or redistribution to
servers or lists, or reuse of any copyrighted component of this work in
other works.}
\thanks{This work was supported by the Office of Naval Research grant N00014-19-1-2556, Army Research Office grant W911NF-18-1-0325 and Princeton University's School of Engineering and Applied Science through the generosity of Lydia and William Addy '82.}
\thanks{$^*$ Equal contribution.}
\thanks{K. Nakamura, M. Santos, and N. E. Leonard are with the Department of Mechanical and
Aerospace Engineering, Princeton University, Princeton, NJ, 08544 USA; 
\href{mailto:k.nakamura@princeton.edu}{\tt\footnotesize k.nakamura@princeton.edu}, \href{mailto:maria.santos@princeton.edu}{\tt\footnotesize maria.santos@princeton.edu}, \href{mailto:naomi@princeton.edu}{\tt\footnotesize naomi@princeton.edu}}}

\begin{document}

\maketitle
\thispagestyle{empty}
\pagestyle{empty}

\begin{abstract}
This paper presents an algorithm for a team of mobile robots to simultaneously learn a spatial field over a domain and spatially distribute themselves to optimally cover it. Drawing from previous approaches that estimate the spatial field through a centralized Gaussian process, this work leverages the spatial structure of the coverage problem and presents a decentralized strategy where samples are aggregated locally by establishing communications through the boundaries of a Voronoi partition. We present an algorithm whereby each robot runs a local Gaussian process calculated from its own measurements and those provided by its Voronoi neighbors, which are incorporated into the individual robot's Gaussian process only if they provide sufficiently novel information. The performance of the algorithm is evaluated in simulation and compared with centralized approaches. 
\end{abstract}

\section{INTRODUCTION}
\label{sec:intro}
The problem of positioning a team of robots over a domain to optimally monitor its environmental features is referred to as the spatial coverage problem. Standard methods that tackle this problem (e.g. \cite{Cortes2004coverage,Schwager06distributedcoverage}) provide control laws for robots to coordinate on their spatial distribution, concentrating around areas of high interest in the domain and spreading out over areas of low interest. 
Depending on the application, the multi-robot team may need to monitor different phenomena, ranging from population density for wireless coverage \cite{Schwager2009decentralized} to radiation levels in the event of a nuclear disaster \cite{schwager2017robust}, or weather conditions in environmental forecasting \cite{Dunbabin2012robots}.

The prevalence of the environmental feature(s) to be monitored over a domain can be  encoded as a spatial field. Frequently referred to as the \textit{density function} in the context of coverage, the spatial field gives each point in the domain a scalar value that represents the relative importance of the point, with higher values of the density corresponding to regions of higher interest. Depending on the robotic application, different functions have been used in the problem of coverage including densities representing static fields \cite{Cortes2004coverage,Schwager06distributedcoverage}, dynamic densities that reflect changes of the field over time \cite{Leonard_etalProcIEEE2007,Pimenta2010simultaneous,Lee2015multirobot,Santos2019decentralized}, collections of densities that represent the presence of heterogeneous events in the environment \cite{Santos2018heterogeneousRAL, Santos2018heterogeneousIROS}, or nonuniform densities that generalize the problem to domains with non-Euclidean distances \cite{LekLeoJCO2009,Bhattacharya2014coverage}. A common assumption in these works, however, is that the multi-robot team knows the density function a priori, and thus is solely concerned with optimally distributing themselves to monitor the domain.

This paper addresses the problem of multi-robot spatial coverage when the density function over the domain is initially unknown. To address this problem, the robots perform two different, albeit complementary, tasks: they take samples needed to estimate the density function and they position themselves in the domain to effectively monitor it. These tasks may be completed sequentially  \cite{schwager2017robust} or concurrently \cite{Benevento2020multi,Santos2021multirobot}. However, while the coverage task can be executed in a decentralized fashion by leveraging the well-known Lloyd's algorithm \cite{Lloyd1982least}, learning the density function typically requires a costly central computation that processes all the samples from the team to produce an estimate \cite{Benevento2020multi,Luo2018}. 
Instead, we propose a fully decentralized approach to simultaneously learn and cover a spatial field. Building on the Voronoi Estimation Coverage (VEC) algorithm from \cite{Santos2021multirobot}, which uses the Voronoi partition to limit the sampling area of each robot but requires a centralized computation to estimate the spatial field, we design a completely decentralized learning method where each robot aggregates its samples and its Voronoi neighbors' samples to produce its own estimate of the density function.  

The paper is organized as follows. This remainder of this section includes a summary of related literature on multi-robot learning for spatial coverage. Section \ref{sec:coverage} introduces the problem of coverage of unknown spatial fields and gives a brief overview of the VEC algorithm in \cite{Santos2021multirobot}. The proposed algorithm, which leverages the Voronoi partition from the coverage objective for the purposes of reducing sampling, is outlined in Section \ref{sec:decentralizedlearning}, with Section \ref{sec:communication} focusing on how data is filtered to reduce the amount of communications needed. Section \ref{sec:simulations} presents a series of simulations that validate the approach and compare it to its centralized counterparts. Conclusions are detailed in Section \ref{sec:conclusions}.

\subsection{Related Work}
\label{sec:literature}
Cooperative estimation of unknown spatial fields has been the subject of study in a variety of multi-robot system applications, e.g.  \cite{Singh2009nonmyopic,ZhaLeoTAC2010,PetersonPaley2013,Hollinger2014sampling,notomista2020communication}. In the context of spatial coverage, different strategies have been proposed to combine optimal coverage of the domain and sampling of the unknown spatial field in the control policy of the robots. 

Some works optimize the robots' path only for the coverage objective, and the sampling locations occur as a direct consequence of the coverage of the current estimate of the field \cite{Luo2018,Luo2019distributed,Benevento2020multi}. Other works opt for a sequential execution of both tasks, typically having the robots estimate the spatial field over the environment and then monitor it accordingly \cite{schwager2017robust,Carron2015multi}.

The simultaneous estimation and coverage of an unknown spatial field has been addressed in \cite{Schwager2009decentralized,Santos2021multirobot}. In 
\cite{Schwager2009decentralized}, the density function is modeled using basis function approximation. However, this approach limits the types of functions that can be learned as it can lead to very large errors or instabilities when the true density function is not well represented by the chosen basis functions. While the authors resolved this instability issue in \cite{schwager2017robust}, the execution of the learning and coverage tasks became sequential. In \cite{Santos2021multirobot} an algorithm was proposed to concurrently learn and cover the spatial field. However, the estimate of the function was modeled as a Gaussian process that aggregates all the samples collected by the robots in the team, and thus, the algorithm relies on a centralized computation. In addition, if the team collects $P$ samples, the Bayesian update of the Gaussian process requires an inversion of a $P\times P$ matrix, which scales poorly with the number of measurements ($\mathcal{O}(P^3)$). In this paper, we maintain the generality afforded by Gaussian processes but leverage the spatial structure given by the Voronoi partition to decentralize the learning objective. We propose a fully decentralized algorithm for simultaneous learning and coverage of spatial fields.

\section{COVERAGE OF UNKNOWN SPATIAL FIELDS}
\label{sec:coverage}
The multi-robot coverage problem deals with how to position a team of mobile robots such that they can optimally monitor the environmental features in a domain \cite{Cortes2004coverage,Schwager06distributedcoverage}. In this context, a resource being more prevalent in an area or an event being more likely may impact the relative importance of  points in the domain to be covered. Differences in the relative importance of  points can be modeled with a density function over the domain that assigns higher values to points that are to be monitored more closely. 

Consider a team of $M$ mobile robots that are to cover a closed and convex domain $D\subset \bbR^d$ (e.g., $d = 2$ for ground robots, $d=3$ for aerial robots) with respect to an initially unknown spatial field $f^*:D\to\mathbb{R}$. 
For the coverage task, a typical approach is to partition the domain into regions of dominance such that Robot $i$, with position $x_i \in D$, is in charge of monitoring the set $V_i$ of points in the domain that are closest to it:
\begin{equation}
\label{eq:voronoi}
    V_i(\bfx) = \{ q \in D ~ | ~ \| q -x_i \| \leq \| q-x_j \|, \;\;  \forall i \neq j \},
\end{equation}
with $\bfx = {[x_1\tr,\dots,x_M\tr]}\tr$ the stacked positions of the robots and $i,j \in \{1,\dots,M\}$. Note that Robot $i$ can compute $V_i$ from the intersection of the bisectors of each segment $x_ix_j$, $j$ a Delaunay neighbor of $i$. Assuming that the sensing radius of each robot is big enough, each robot can compute its Delaunay neighbors locally, e.g. as in \cite{Guibas1992Delaunay}.



With this partition, the quality of  coverage of  Robot $i$ with respect to the density function $f^*$ is inversely related to cost
\begin{equation}
\label{eq:cost}
    h_i(\bfx, f^*(\cdot)) = \int_{V_i(\bfx)} \lVert q - x_i \rVert^2 f^*(q)dq.
\end{equation}
The squared distance models the decaying quality of sensor data at point $q$ with increasing distance from the robot at $x_i$. The performance of the multi-robot team can be evaluated as the sum of the individual robot costs:
\begin{equation}
\label{eq:totalcost}
    H(\bfx,f^*(\cdot)) = \sum_{i=1}^M h_i(\bfx, f(\cdot)) = \sum_{i=1}^M \int_{V_i(\bfx)} \lVert q - x_i \rVert^2 f^*(q)dq,
\end{equation}
with a better coverage corresponding to lower values of $H$. 

Finding a global minimum to \eqref{eq:totalcost} is an NP-hard problem \cite{Luo2019distributed}. However, one configuration that is known to locally minimize the cost \eqref{eq:totalcost} is the Centroidal Voronoi Tesselation (CVT) \cite{Du1999cvt}. In a CVT, the position of each robot corresponds to the center of mass of its Voronoi cell, given by
\begin{equation}
\label{eq:cm}
    c_i(\bfx,f^*(\cdot)) = \frac{\int_{V_i(\bfx)} q f^*(q)dq}{\int_{V_i(\bfx)} f^*(q)dq}.
\end{equation}
The multi-robot team can thus minimize \eqref{eq:totalcost} by making each robot move towards its center of mass \cite{Cortes2004coverage}, which corresponds to a continuous version of Lloyd's algorithm \cite{Lloyd1982least}, \begin{equation}\label{eq:lloyds}
    \dot x_i = \kappa (c_i(\bfx,f^*(\cdot)) - x_i), \quad \kappa > 0, \quad \forall i\in\{1,\dots,M\}.
\end{equation}


A strong assumption behind the control law in \eqref{eq:lloyds} is that each robot knows the spatial field $f^*(\cdot)$ and, thus, is able to compute the center of mass in \eqref{eq:cm}. The recently developed Voronoi Estimation Coverage (VEC) algorithm \cite{Santos2021multirobot} leverages Gaussian process regression to learn the estimate of the true spatial field. The estimate $f(\cdot)$ of the true function $f^*(\cdot)$ is  modeled as a Gaussian process
\begin{equation}\label{eq:estimateoff}
    f(\cdot) \sim GP(\mu(\cdot\: ; \rho), k(\cdot, \cdot \: ; \tau)),
\end{equation}
with $\mu:D\to\mathbb{R}$ and $k:D\times D\to\mathbb{R}$ the mean and covariance functions, respectively, and $\rho$ and $\tau$, the hyperparameters. Adopting a linear mean and a squared-exponential covariance function as in \cite{Santos2021multirobot},  we have
\begin{equation}
\label{eq:muAndKernel}
    \mu(x;\rho) = \rho\tr x,  \;\;\;\;\; k(x,x',;) = \text{exp}\left(-\frac{\lVert x - x' \rVert ^2}{2\tau^2}\right).
\end{equation}

The mean and covariance in \eqref{eq:muAndKernel}, characterized by  $\rho$ and $\tau$,  can be approximated by maximum-likelihood estimation from noisy measurements of the density function \cite{rasmussen2003gaussian}. 
Robot $i$ takes a noisy scalar measurement $y_i^{(n)}$ at iteration $n$:
\begin{equation}
\label{eq:samp}
\begin{aligned}
    y_i^{(n)} &= f^*(x_i) + \epsilon, 
\end{aligned}
\end{equation}
with $\epsilon \sim \mathcal{N}(0,\sigma_m^2)$, where $\sigma_m^2$ models the variance in the measurement. Denote by $N$ the total number of iterations, with $n\in\{1,\dots N\}$ the time iteration. The set of points sampled in the domain $D$ at time $n$ is denoted by $\bfx^{(n)} = \left[{x_1^{(n)}}^\intercal, \dots, {x_M^{(n)}}^\intercal\right]^\intercal \in \mathbb{R}^{Md}$ and the corresponding  measurements are $\bfy^{(n)}=[y_1^{(n)},\dots,y_N^{(n)}]^\intercal\in\bbR^M$. We let $\bfx^{(1:n)}$ and $\bfy^{(1:n)}$ refer to all the sampled locations and measurements taken from iteration 1 to $n$.

The centralized maximum likelihood estimate for a Gaussian process is known in closed form (see, e.g. \cite{rasmussen2003gaussian}) as
\begin{equation}
\label{eq:maxLikelihood}
\begin{split}
   & (\rho^{(n)} , \tau^{(n)}) =  \argmin_{\rho, \tau} \big\{ \log|\mathbf{K}_{\mathbf{x}^{(1:n)}}\!(\tau) + \sigma_m^2\mathbf{I}| + \\
    & 
    {(\pmb{\mu}^{(1:n)}\!(\rho) \!-\! \mathbf{y}^{(1:n)})\!}^\intercal 
    (\mathbf{K}_{\mathbf{x}^{(1:n)}}\!(\tau) \!+\! {\sigma_m^2\mathbf{I})\!}^{-1} 
    (\pmb{\mu}^{(1:n)}\!(\rho) \!-\! \mathbf{y}^{(1:n)\!}) \big\},
\end{split}
\end{equation}
with $\bfI$ the $Mn \times Mn$ identity matrix, and $\bfK_{\bfx^{(1:n)}}(\tau)\in\bbR^{Mn\times Mn}$ the covariance matrix with entries $ k(x_{i}^{(j)},x_{i'}^{(j')};\tau)$ for $i,i'\in\left\{1,\dots, M\right\}$ and $j,j'\in\left\{1,\dots, n\right\}$. Finally, $\pmb{\mu}^{(1:n)}(\rho)$ represents the mean output of the GP.

With the estimate of the hyperparameters, the robot team reconstructs the mean and covariance functions as  
\begin{align}
&    \mu^{(n)}(x) = \mu(x;\rho^{(n)}) + \\
    &\bfk(x;\mathbf{x}^{(1:n)})^\intercal(\mathbf{K}_{\mathbf{x}^{(1:n)}}(\tau^{(n)}) \!+\! \sigma_m^2\mathbf{I})^{-1} 
    (\mathbf{y}^{(1:n)} \!-\! \pmb{\mu}^{(1:n)}(\rho^{(n)})),\\
 &   k^{(n)}(x, x') = k(x,x';\tau^{(n)}) - \\ &\bfk(x;\mathbf{x}^{(1:n)})^\intercal(\mathbf{K}_{\mathbf{x}^{(1:n)}}(\tau^{(n)}) + \sigma_m^2\mathbf{I})^{-1} 
    \bfk(x';\mathbf{x}^{(1:n)}),
    \label{eq:paramK}
\end{align}
where $\bfk(x;\bfx^{(1:n)})\in\mathbb{R}^{Mn}$ is the vector with entries $k(x_i^{(j)},x;\tau^{(n)})$ for $i\in\{1,\dots,M\}$ and $j\in\{1,\dots,n\}$. 
The reconstructed covariance function is then used to estimate the standard deviation 
\begin{equation}
    \sigma^{(n)}(x) = \sqrt{k^{(n)}(x,x)}. \label{eq:paramSigma}
\end{equation}

Note the inversion of the matrix $(\mathbf{K}_{\mathbf{x}^{(1:n)}}(\tau) + \sigma^2\mathbf{I})$ in \eqref{eq:maxLikelihood}. The matrix $\mathbf{K}_{\mathbf{x}^{(1:n)}}$ contains the kernel evaluation \eqref{eq:muAndKernel} between every pair $(x_{i}^j, x_{i'}^{j'})$, elements of $\mathbf{x}^{(1:n)}$ and is size $Mn \times Mn$. This computation requires each agent to send information to a centralized computer at each time step to perform this evaluation and scales $\mathcal{O}(M^3n^3)$.

At each time step $n \in \{1,\dots, N\}$, the VEC algorithm updates the maximum likelihood estimate of $\rho^{(n)}$ and $\tau^{(n)}$.  These are used to create an estimate of the density function as in \cite{Benevento2020multi,Santos2021multirobot},
\begin{equation}
\label{eq:vec_robo_est}
    f^{(n)}(q) = \mu^{(n-1)}(q) - \sqrt{\beta^{(n)}}\sigma^{(n-1)}(q), \quad q\in D.
\end{equation}
Here, $\{\beta^{(n)}\}_{n=1}^N$ is a non-negative non-decreasing sequence that leads to a small underestimate of the prediction. This encourages robots to be ``pushed'' to regions of higher importance, i.e. where the mean of estimate is higher \cite{Benevento2020multi}. Using this estimate, each robot selects a goal point which is a linear interpolation between the center of mass $c_i^{(n)}$ and a point $e_i^{(n)}$ that maximizes the uncertainty in \eqref{eq:paramSigma} within its respective Voronoi cell. Regarding the center of mass, $c_i^{(n)}$, note: (i) $c_i^{(n)}$ is computed using \eqref{eq:cm} with the estimate  $f^{(n)}(\cdot)$, and, thus, minimizes $H(\cdot,f^{(n)})$;  and (ii), during iteration $n$, $c_i^{(n)}$ is not a static point---the center of mass evolves as a consequence of the movement the robots, which impacts the Voronoi partition of the domain.
The controller thus  sends each robot to their goal point:
\begin{equation}
\label{eq:controller}
    \dot x_i = \kappa\left( (1-\gamma^{(n)})c_i^{(n)} + \gamma^{(n)}e_i^{(n)} - x_i\right), \;\; \kappa > 0.
\end{equation}
The weighting of the linear interpolation is determined by  $\{\gamma^{(n)}\}^N_{n=1}$, a nonnegative decreasing sequence less than 1.  The weighting between the two points varies between timesteps, such that at later times more preference is given to points that minimize the cost (\ref{eq:totalcost}). This linear interpolation is guaranteed to be within the domain and remain within each robot's respective Voronoi cell due to the assumed convexity of the domain. When the equilibrium of \eqref{eq:controller} is achieved throughout the team, each robot collects its noisy sample of the environment, $y_i^{(n)}$. These are centrally aggregated to update the maximum likelihood estimate of the hyperparameters $\rho$ and $\tau$ as in \eqref{eq:maxLikelihood}, which leads to a new iteration of the algorithm.

The VEC algorithm has shown remarkable performance for the problem of simultaneously learning and covering a spatial field \cite{Santos2021multirobot}. 
One large drawback of the method, however, is the centralized computation of the hyperparameters $\rho$ and $\tau$. For a large team of robots collecting samples over a long time horizon, this computation can become prohibitively expensive as samples are continuously collected and also leaves the system vulnerable should anything happen to the centralized computer. 

\section{DECENTRALIZED LEARNING WITH VORONOI COMMUNICATIONS}
\label{sec:decentralizedlearning}
The main contribution of this paper is an extension to the VEC algorithm that decentralizes the computation of the hyperparameters $\rho$ and $\tau$. Our proposed solution is motivated by the fact that the decision making of Robot $i$ when computing $c_i^{(n)}$ and $e_i^{(n)}$ 
for the controller in \eqref{eq:controller} 
depends only on the estimate $f^{(n)}(\cdot)$ and uncertainty $\sigma^{(n)}(\cdot)$ within its cell $V_i$ and its vicinity. The information far from the robot has no impact. By focusing only on the samples collected by a robot and those of its adjacent neighbors, a robot will still be able to recover an accurate estimate of the density function in its vicinity without incurring prohibitive computational costs.

We propose the Decentralized Voronoi Estimation Coverage (DVEC) algorithm (see Algorithm \ref{alg:DVEC}), where each robot has an individual estimate of the density function, $f_i(\cdot)$. This estimate is modeled as a Gaussian process regression,
\begin{equation}
    f_i(\cdot) \sim GP(\mu_i(\cdot\: ; \rho_i), k_i(\cdot, \cdot \: ; \tau_i)),
\end{equation}
with the mean $\mu_i$ and covariance $k_i$ functions having the same structure as in \eqref{eq:muAndKernel}. The novelty of this approach relies on the fact that each robot calculates this regression only with its own samples and samples received from its neighbors. 

In this section we introduce two communication protocols that regulate the information exchange between neighboring robots, which we refer to as \textit{naive} and \textit{constrained communication}. In the first case, robots always exchange samples with their Delaunay neighbors. In the second, we introduce an energy-saving criterion that allows robots to discard samples that do not add substantial information to the density model, thus reducing the computational burden on the regression. In both cases there is an implicit assumption that communication for this is synchronous. This is due to our algorithm not allowing an individual robot to take a measurement until the entire team  has converged to the equilibrium of \eqref{eq:controller}.

\subsection{Naive Communication Protocol}
\label{sec:naive}
The DVEC algorithm takes advantage of the Voronoi tessellation \eqref{eq:voronoi} to create a communication network between the robots where they can exchange information about their samples of the environment. Using the spatial structure given by the partition, a naive way of aggregating samples for the Gaussian process is to let Robot $i$ exchange information about its sample $\{x_i^{(n)},y_i^{(n)}\}$ with its Delaunay neighbors. Let $\mathcal{N}_i$ denote the set of neighbors of Robot $i$ in the Voronoi partition, i.e. all robots $j$ whose Voronoi cell $V_j$ shares a boundary with $V_i$. If all robots share their samples with their Voronoi neighbors, the information available to Robot $i$ at iteration $n$ can be written as the sets,
\begin{equation}
\label{eq:naiveCom}
\begin{split}
    &\bfx_i^{(n)} = \{x_i^{(n)}\} \:\cup \: (\bigcup_{j \in \mathcal{N}_i} \:\{x_j^{(n)}\},\\
    &\bfy_i^{(n)} = \{y_i^{(n)}\} \: \cup \: (\bigcup_{j \in \mathcal{N}_i} \:\{y_j^{(n)}\}).
\end{split}
\end{equation}
In a slight abuse of notation, we use the symbols $\bfx_i^{(n)}$ and $\bfy_i^{(n)}$ to indistinctly denote the sets in \eqref{eq:naiveCom} and the column vectors whose entries correspond to the stacked elements of those sets (analogously to the definition of $\bfx^{(n)}$). The ordering of the elements in $\bfx_i$ and $\bfy_i$ does not affect the computation of the Gaussian process regression. Furthermore, we denote as $\bfx_i^{(1:n)}$ and $\bfy_i^{(1:n)}$ all the samples aggregated up to iteration $n$. Note that the neighbors of Robot $i$ at each iteration may vary: in general,  $\mathcal{N}_i^{(l)}\neq\mathcal{N}_i^{(l')}$ for two different iterations $l,l'$.

After communicating with its neighbors, Robot $i$ computes its own estimate of the hyperparameters $\rho_i$ and $\tau_i$ as in \eqref{eq:maxLikelihood} using in this case $\bfx_i^{(1:n)}$ and $\bfy_i^{(1:n)}$. This allows Robot $i$ to maintain its own estimate of the spatial field $f_i(\cdot)$ and the corresponding uncertainty $\sigma_i(\cdot)$, calculated as in \eqref{eq:vec_robo_est} and \eqref{eq:paramSigma}, respectively. The controller then interpolates its goal point between the center of mass calculated with respect to its individual estimate, $c_i^{(n)}$, and the point that maximizes the uncertainty in the cell $e_i^{(n)}$. The pseudocode of the algorithm for an individual robot is outlined in Algorithm \ref{alg:DVEC}. The algorithm follows closely the steps of VEC, 
but DVEC considers all the learning computations at the individual level. 

The naive communication strategy adopted in DVEC results in each robot having an estimate calculated based on the information gathered in its vicinity, but that has less data and, as a result, more uncertainty in regions of the domain that are far away from the robot. However, in most cases, the most relevant information for a robot to maximize its contribution to the coverage objective relates to the estimate of the density within the robot's Voronoi cell and its immediate surroundings. Thus, we do not anticipate severe performance losses by using a local estimate of the spatial field for the coverage task. In Section \ref{sec:simulations} we verify this assumption in a series of simulations.

\begin{algorithm}[!t] \label{alg:DVEC}
\caption{Decentralized Voronoi Estimation Coverage (DVEC)}
 \SetAlgoLined
    \tcc{Algorithm for Robot $i$}
 
    \KwIn{Parameter $\kappa,$ Domain $D,$ sequences $\{\beta^{(n)}\}_{n=1}^N$ and  $\{\gamma^{(n)}\}_{n=1}^N$}
    {\bf{Initialization\;}}
    Random initial position $x_i^{(0)}$
    
    Sample initial position $y_i^{(0)}= f^* (x_i^{(0)})+\epsilon_i^{(0)}$
    
    Exchange data $\{ \bfx_i^{(0)}, \bfy_i^{(0)} \}$
    
    
    
    Calculate parameters $\rho_i^{(0)}$ and $\tau_i^{(0)}$
        
    Calculate mean $\mu_i^{(0)}(\cdot)$ and standard deviation $\sigma_i^{(0)}(\cdot)$
    
    \For{\text{\normalfont\ each iteration}  $n\in \left\{1,\dots, N\right\}$}{
    
            Calculate the estimate of $f^*$ for all $q\in D$ as
            $
                ~~f_i^{(n)}(q)=\mu_i^{(n-1)}(q)-\sqrt{\beta^{(n)}}\sigma_i^{(n-1)}(q)
            $
        Choose estimation position as
            $~~e_i^{(n)}=\argmax_{x\in {V}_i^{(n-1)}}\sigma^{(n-1)}(x)$
            
        \While{$x_i \neq (1-\gamma)c_i^{(n)} + \gamma e_i^{(n)}$}{
            Compute $c_i^{(n)}(\cdot,f_i^{(n)}(\cdot))$ as in \eqref{eq:cm}
            
            Move according to $\dot x_i=\kappa\left((1-\gamma^{(n)})c_i^{(n)}+\gamma^{(n)}e_i^{(n)}-x_i\right)$       
        }

            Sample $y^{(n)}_i = f^* (x_i^{(n)})+\epsilon_i^{(n)}$
            
            Exchange data $\{ \bfx_i^{(n)}, \bfy_i^{(n)} \}$
            
            Update $\rho_i^{(n)}$ and $\tau_i^{(n)}$
            
            Update $\mu_i^{(n)}(\cdot)$ and $\sigma_i^{(n)}(\cdot)$
            
        }

\end{algorithm}





\begin{figure*}
  \subfloat[\label{subfig:a}]{\includegraphics[width=.24\textwidth]{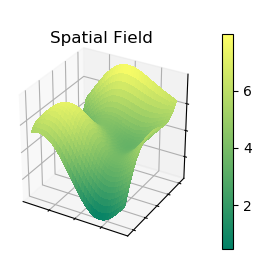}}
  \subfloat[\label{subfig:b}]{\includegraphics[width=.24\textwidth]{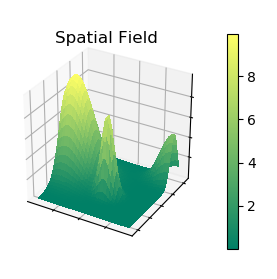}}
  \subfloat[\label{subfig:c}]{\includegraphics[width=.24\textwidth]{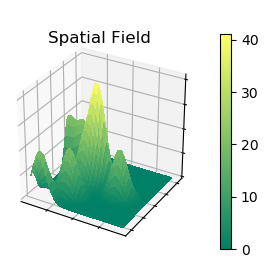}}
  \subfloat[\label{subfig:d}]{\includegraphics[width=.24\textwidth]{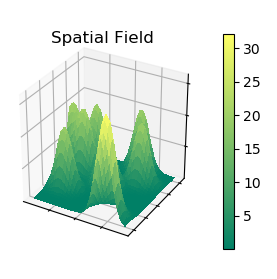}}
  \caption{Density functions used in the simulations. \protect\subref{subfig:a} shows the density function from \cite{Benevento2020multi}. The densities in \protect\subref{subfig:b}, \protect\subref{subfig:c}, and \protect\subref{subfig:d} are from \cite{Santos2021multirobot}: \protect\subref{subfig:b} shows a sum of 3 Gaussian bivariate distributions with different means and variances, and the densities in \protect\subref{subfig:c} and \protect\subref{subfig:d} are sums of 9 Gaussian bivariate distributions with equal variances generated with random means.}
  \label{fig:densities}
\end{figure*}

\subsection{Constrained Communication Protocol}
\label{sec:communication}
The communication protocol in Section \ref{sec:naive} considers a scenario when agents always incorporate their neighbors' samples to their model, irrespectively of the value added by such information. In this section, we present an alternative protocol that limits the amount of information shared between the agents in an effort to reduce redundant or uninformative transmissions that overload the communication system and increase the number of computations in the regression process.


Consider the collection of samples of Robot $i$, $\bfy_i^{(1:n)}$ at locations $\mathbf{x}_i^{(1:n)}$. We would like to know if a new sample $y_j^{(n)}$ obtained by Robot $j$, $j\in\mathcal{N}_i$ should be added to $\bfx_i^{(n)}$ by analyzing the amount of potential information that $\{x_j^{(n)},y_j^{(n)}\}$ would add to the estimate $f_i$. To this end, the communication constrained estimation algorithm in \cite{notomista2020communication} provides an upper bound on the difference between the prediction of the spatial field $f_i(\bar x)$ at a location $\bar x \in D$ with and without incorporating a new sample at a location $\chi_j \in D$ to the Gaussian process regression. This is formally notated as 
\begin{equation}
    | f_i(\bar x ; \mathbf{x}_i^{(1:n)}) - f_i(\bar x ; \mathbf{\tilde{x}}_i^{(1:n)})|,
\end{equation}
where $\mathbf{\tilde{x}}_i^{(1:n)} = \mathbf{x}_i^{(1:n)} \cup \chi_j $. We analogously use the notation $\mathbf{\tilde{y}}_i^{(1:n)} = \mathbf{y}_i^{(1:n)} \cup \psi_j$, with $\psi_j$ is the noisy sample taken at $\chi_j$ according to \eqref{eq:samp}. The notation $f_i(\bar x;\bfx_i^{(1:n)})$ is used here to indicate that the estimate of the density function at $\bar x$ is obtained by using the collection of samples associated with $\mathbf{x}_i^{(1:n)}$ in the Gaussian process regression of $f_i$.

The bound is 
\begin{equation}
    | f_i(\bar x ; \mathbf{x}_i^{(1:n)}) - f_i(\bar x ; \mathbf{\tilde{x}}_i^{(1:n)})| \leq \lVert \bfk(\bar x ; \mathbf{\tilde{x}}_i^{(1:n)})^\intercal \rVert \lVert \mathbf{\tilde{y}}_i^{(1:n)} \rVert \Delta K,
\end{equation}
where
\begin{equation}
\label{deltak}
    \Delta K \leq \max \{ \|\mathcal{X} \|, \| \mathcal{Z}\| \} + \|\mathcal{Y}\|,
\end{equation}
and 
\begin{equation}
\label{xyz}
\begin{aligned}
    &\mathcal{Z} = \begin{aligned}[t]  \big(&k(\chi_j, \chi_j; \tau_j) - \\
    &\mathbf{k}(\chi_j ; \mathbf{x}_i^{(1:n)})^{-1}\mathbf{K}_{\mathbf{x}_i^{(1:n)}}(\tau_i)^{-1}\mathbf{k}(\chi_j ; \mathbf{x}_i^{(1:n)})\big)^{-1} \end{aligned}\\    &\mathcal{Y} = -\mathbf{K}_{\mathbf{x}_i^{(1:n)}}(\tau_i)^{-1}\mathbf{k}(\chi_j ; \mathbf{x}_i^{(1:n)}) \; \mathcal{Z} \\
    &\mathcal{X} = -\mathcal{Y}\;{\mathbf{k}(\chi_j ; \mathbf{x}_i^{(1:n)})}^{\intercal} \mathbf{K}_{\mathbf{x}_i^{(1:n)}}(\tau_i)^{-1}
\end{aligned}
\end{equation}
Bounds on these terms can be evaluated using the Cauchy-Schwarz inequality and the feature-space view of the kernel function as in \cite{notomista2020communication},
\begin{equation}
\label{bounds}
\begin{aligned}
    &\|\mathcal{X}\| \leq \| \mathbf{K}_{\mathbf{x}_i^{(1:n)}}(\tau_i)^{-1}\mathbf{k}(\chi_j ; \mathbf{x}_i^{(1:n)})\|^2 \|\mathcal{Z}\|, \\
    &\|\mathcal{Y}\| \leq \| \mathbf{K}_{\mathbf{x}_i^{(1:n)}}(\tau_i)^{-1}\mathbf{k}(\chi_j ; \mathbf{x}_i^{(1:n)})\|\|\mathcal{Z}\|, \\
        &\|\mathcal{Z}\| \leq \frac{|\frac{1}{k(\chi_j, \chi_j; \tau_j)}| + (\frac{\|\mathbf{k}(\chi_j ; \mathbf{x}_i^{(1:n)})\|}{k(\chi_j, \chi_j; \tau_j)})^2\|\mathbf{K}_{\mathbf{x}_i^{(1:n)}}(\tau_i)^{-1}\|}{1 - (\frac{\|\mathbf{k}(\chi_j ; \mathbf{x}_i^{(1:n)})\|}{k(\chi_j, \chi_j; \tau_j)}\|\mathbf{K}_{\mathbf{x}_i^{(1:n)}}(\tau_i)^{-1}\|)^2}.
\end{aligned}
\end{equation}

The notation above is in principle the same as from Section \ref{sec:coverage}; however, the dimensions of the matrix $\mathbf{K}_{\mathbf{x}_i^{(1:n)}}(\tau_i)$ and vector $\mathbf{k}(\chi_j ; \mathbf{x}_i^{(1:n)})$ cannot be stated \textit{a priori}. Their dimensions depend on the communications between agent $i$ and its neighbors up until iteration $n$.
In \eqref{xyz}, $\mathcal{Z}$ represents the covariance in the prediction at the test point given our training points in the first $n$ iterations. The other factors contributing to $\mathcal{X}$ and $\mathcal{Y}$ are related to the expression for the mean predictions at a test point \cite{rasmussen2003gaussian}. A more detailed treatment can be found in \cite{notomista2020communication}.

Using the method outlined in \cite{notomista2020communication}, one can quantify the impact a new data point will have on the estimate based on the value of $\Delta K$, which is computed using \eqref{bounds}. \cite{notomista2020communication} used this value to let a robot rank the impact of every candidate data point to be received and accepted only the most impactful points from other robots within a positive radius $R$ from itself. In contrast, our implementation sets a threshold value for $\Delta K$ and determines if a sample from a neighbor at iteration $n$, $\{x_j^{(n)},y_j^{(n)}\}$, $j\in\mathcal{N}_i^{(n)}$ should be added or not to the Gaussian process of $f_i^{(n)}$ based on the value of $\Delta K$ associated with the sample. 

\section{SIMULATION RESULTS}
\label{sec:simulations}
The algorithms DVEC with naive and constrained communications (henceforth referred to as DVEC-nc and DVEC-cc, respectively) were compared to the baseline, VEC, in a series of simulations to determine the impact of considering local estimates of the density function. All algorithms were tested in a collection of four different spatial fields, shown in Fig. \ref{fig:densities}. 
Each trial ran for 15 iterations with a team of $M=7$ robots, each one moving according to single integrator dynamics, i.e., $\dot x_i = u, \; u\in\mathbb{R}$ given by the VEC and DVEC algorithms.

In order to compare the performance of VEC, DVEC-nc and DVEC-cc, we use the error in the estimate of the density function as defined in \cite{Santos2021multirobot}, 
\begin{equation}
    \Delta^{(n)} = \int_D |f^{(n)}(q) - f^{*}(q)|dq.
\label{err_eq}
\end{equation}
However, while in the case of VEC the estimate of the function is uniquely defined at iteration $n$ by $f^{(n)}$, for DVEC we have a collection of estimates $\{f_i^{(n)}\}_{i=1}^M$. In order to directly compare against a the centralized estimation of VEC, a composition method from \cite{Santos2018heterogeneousRAL} was used to combine the estimates for analysis at each iteration. 
For this comparison, we use the composition defined by taking the spatial field estimate that gives the maximum value of the density $\max_{i} f_i^{(n)}(q), \forall q \in D$. 
Since the estimate of the spatial field given by \eqref{eq:vec_robo_est} underestimates the mean estimate of the field proportionally with the uncertainty of the measurement at that point, taking the maximum was shown in testing to lead to lower errors on the composed density estimate as compared to using other composition operators. The composition can be parallelized as each point in the discretization of the domain is independent of the others.

\begin{figure}[t]
  \centering
  \includegraphics[width=1\columnwidth]{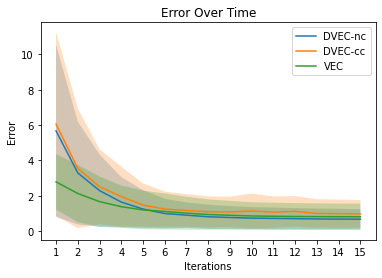}
  \caption{Error in estimate as calculated by \eqref{err_eq} for VEC (green), DVEC-nc (blue) and DVEC-cc (orange) over 15 iterations from 5 different initial conditions of the multi-robot team. The shaded regions represent  standard deviation across the trials over the four tested densities in Fig. \ref{fig:densities}.} 
\label{error_time}
\end{figure}

The comparison on the error in the density estimate is shown in Fig. \ref{error_time}. The error in the DVEC algorithms is initially higher than VEC but quickly reaches a similar level by iteration 6 for both communication methods. Afterwards, 
the estimate ends up with near identical mean error (0.815 for VEC, 0.678 for DVEC-nc, and 0.888 for DVEC-cc at iteration 15). The larger standard deviation for the DVEC experiments stem from the initial communications between agents; since communication is only allowed between Voronoi neighbors, the initial configuration impacts the initial estimate much more than if the estimates were computed centrally. The standard deviations at the final iteration were similar: 0.743 for VEC, 0.568 for DVEC-nc, and 0.681 for DVEC-cc.

\begin{figure}[t]
  \centering
  \includegraphics[width=1\columnwidth]{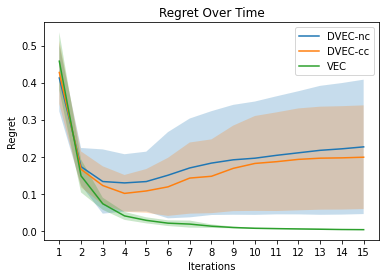}
  \caption{Regret as calculated by \eqref{regret_eq} for VEC (green), DVEC-nc (blue) and DVEC-cc (orange) over 15 iterations from 5 initial conditions. The shaded regions represent  standard deviation across the trials over the four tested densities.}
  \label{regret_time}

\end{figure}

The algorithms are also compared in terms of regret at each iteration, defined as
\begin{equation}
    r^{(n)} = {H}(f^*(\cdot), \mathbf{x}^{(n)}) - \min_x {H}(f^*(\cdot), \mathbf{x}).  
\label{regret_eq}
\end{equation}
The comparison between the regret for VEC, DVEC-nc and DVEC-cc is shown in Fig. \ref{regret_time}. Here, the regret starts at a similar value for all algorithms since it is a function of the position and true density function, and not the density estimates. At later iterations, the regret between DVEC-nc and DVEC-cc is similar, but larger than VEC. This is likely due to the CVT being only a \textit{local} minimum of the cost function \eqref{eq:totalcost}. Although the estimate $f^{(n)}(\cdot)$ is shown in Fig. \ref{error_time} to converge to the true density $f^*(\cdot)$ for all methods, a better spatial configuration appears  to be established earlier for VEC, around iterations 2 and 3. Since both methods prioritize exploration in early iterations and focus more on coverage later, the configurations do not change much after the first few iterations.

It is important to note that Fig. \ref{error_time} represents composed estimates for DVEC-nc and DVEC-cc, which in this case is $\max_{i} f_i^{(n)}(q), \forall q \in D$. Even though the composed estimate of the robot team may have low error (Fig. \ref{error_time}), the composed estimate does not impact the decisions for individual agents.
 The larger standard deviation and slight increase in regret for the DVEC methods is hypothesized to be due to the incomplete information at the individual level, which may lead to suboptimal coverage positions.  

\begin{figure}[t]
  \centering
  \includegraphics[width=1\columnwidth]{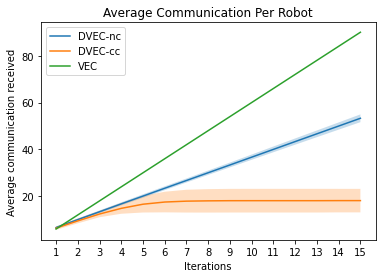}
  \caption{Average accumulated number of information transfers for an individual robot in the 7 robot team over each iteration for VEC (green), DVEC-nc (blue) and DVEC-cc (orange). An information transfer is defined as an incident where a robot transmits a location and sample to another robot.}
  \label{comm_time}

\end{figure}

 The average accumulated number of information transfers per robot is compared between DVEC-nc and DVEC-cc in Fig. \ref{comm_time}. The equivalent number of information transfers to recover the performance from VEC is also shown in green. The plot shows that as iterations increase, the communication constrained protocol DVEC-cc results in fewer transmissions of data across the entire robot team. Both versions of DVEC transfer significantly less information than VEC. 
 The transfer of data for DVEC-cc stops almost entirely around iteration 7, consistent with Fig. \ref{error_time} when the error in density estimate has converged. This is also determined by the thresholding parameter for $\Delta K$. In spite of this reduced communication for DVEC-cc, the regret and error for DVEC-nc and DVEC-cc have experimentally been shown to be nearly identical across a range of spatial fields and initial configurations.  
 
\section{CONCLUSIONS}
\label{sec:conclusions}
In this paper, we introduced a decentralized algorithm for a team of robots to simultaneously learn and cover an unknown spatial field. Leveraging the spatial structure of the coverage problem where individual robots monitor events within their Voronoi cell, we propose a decentralized approach to learning and covering the spatial field whereby each robot runs a Gaussian process that aggregates the samples of itself and its Voronoi neighbors.

Simulations demonstrated the effectiveness of Decentralized Voronoi Estimation Coverage (DVEC) with both naive and constrained communication exchanges between Voronoi neighbors. Communications and computation were reduced by taking into account the novelty of the information being received, rejecting data points that do not greatly improve the estimate of the spatial field. Despite the constrained communication drastically lowering the amount of information transferred between agents, the estimation error and coverage regret was shown to be comparable to DVEC under a naive communication scheme that exchanged information at every step. Both implementations of DVEC presented minor drawbacks in terms of performance when compared to a centralized computer. 



\section{ACKNOWLEDGEMENTS}
The authors thank Professor Gennaro Notomista for his invaluable explanations and insights about communication-constrained coverage. 

\addtolength{\textheight}{-2cm}   






\bibliographystyle{IEEEtran}
\bibliography{biblio}

\end{document}